\documentclass{article}
\usepackage{spconf,amsmath,graphicx,formatfig,xcolor,url}

\title {Robust Super-resolution GAN, with Manifold-Based and Perception Loss}

\name {Uddeshya Upadhyay, Suyash P. Awate\thanks{We thank support from the Nvidia GPU Grant Program, IIT Bombay Seed Grant 14IRCCSG010, and Aira Matrix.}}

\address {Computer Science and Engineering (CSE) Department, Indian Institute of Technology (IIT) Bombay.}

\begin{document}

\maketitle

\begin{abstract}

  Super-resolution using deep neural networks typically relies on highly curated training sets that are often unavailable in clinical deployment
  scenarios. Using loss functions that assume Gaussian-distributed residuals makes the learning sensitive to corruptions in clinical training sets.
  We propose novel loss functions that are {\em robust} to corruptions in training sets by modeling {\em heavy-tailed non-Gaussian} distributions on
  the residuals.
  We propose a loss based on an autoencoder-based {\em manifold-distance} between the super-resolved and high-resolution images, to reproduce
  realistic {\em textural content} in super-resolved images.
  We propose to learn to super-resolve images to match human perceptions of structure, luminance, and contrast.
  Results on a large clinical dataset shows the advantages of each of our contributions, where our framework improves over the state of the art.

  
\end{abstract}

\section {Introduction and Related Work}
\label {sec:intro}

Image super-resolution methods generate high-resolution~(HR) images from acquired low resolution~(LR) images. In many applications, e.g., microscope
slide scanning and digital pathology, acquiring LR images can be much faster than acquiring HR images. LR scanning equipment is also cheaper. So,
methods that generate high-quality super-resolved~(SR) images from LR images can improve productivity in clinical and scientific applications, with
much reduced cost.

Learning based approaches have been effective for super-resolution, e.g., early methods using manifold learning~\cite{sr} and
later methods relying on sparse models~\cite{Yang2010,Mousavi2017} and random forests~\cite{Schulter2015}.
Recent methods leverage deep neural networks (DNNs), e.g., convolutional networks~\cite{Dong2014,Bruna2016} and Laplacian-pyramid
networks~\cite{Lai2017}. DNNs effectively model the rich textural details in HR images and learn complex mapping functions from LR image patches to
the corresponding HR patches.
A state-of-the-art DNN in super-resolution relies on a generative adversarial network (GAN), namely SRGAN~\cite{srgan}. GANs improve texture modeling
in HR patches, and the associated LR-to-HR mapping functions, by using a discriminator in their architecture. During learning, while the discriminator
adapts to separate SR images from HR images, the generator adapts to challenge the discriminator by producing SR images that are progressively closer
to the HR images.

DNN-based super-resolution~\cite{Dong2014,Bruna2016,Lai2017,srgan} relies on high-quality highly curated training sets that entail significant expert
supervision (time, effort, cost). In clinical deployment, training sets is far from perfect because of inherent errors in tissue slicing (e.g.,
tearing), staining (e.g., varying dye concentration), imaging artifacts (e.g., poor focus and contrast, noise), and human mislabeling of data (e.g.,
mixing images from one organ or imaging modality into another).
Typical DNN training uses mean-squared-error (MSE) loss that can make the training very sensitive to (even a small fraction of) atypical, or largely
clinically irrelevant, examples present in the training set. MSE-based losses can force the learning, undesirably, to adapt to atypical examples at
the cost of the performance on the clinically relevant ones.
Thus, we propose novel {\em quasi-norm} based losses that are {\em robust} to errors in dataset curation by modeling {\em heavy-tailed non-Gaussian}
probability density function (PDFs) on the residuals.

In addition to penalizing pixel-wise differences, independently, between SR and HR images, we propose to (i)~learn the manifold of HR images using an
{\em autoencoder} and (ii)~penalize the {\em manifold-distance} between SR and HR images, to capture dissimilarities in {\em textural content}
(factoring out noise and some artifacts) indicated by inter-pixel dependencies.
SRGAN~\cite{srgan} focuses on natural images and uses a VGG-encoding \cite{vgg_net} based loss to learn to generate SR images perceptually similar to
HR images. For histopathology images, we show that the VGG encoder captures human perception sub-optimally.
We propose a loss based on the structural similarity index (SSIM), at a fixed scale, between SR and HR images, to generate SR images that match human
perceptions of structure, luminance, and contrast.

We propose a novel GAN-based learning framework for super-resolution that is {\em robust} to errors in training-set curation, using {\em quasi-norm}
based loss functions.
In addition to independent pixel-wise losses, we propose to learn the {\em manifold} capturing textural characteristics in HR images, by first
(pre-)training an autoencoder and then using a robust penalty on dissimilarities between {\em encodings} of the SR and HR images.
We propose to train our GAN to increase the {\em SSIM} (at a fixed scale) to make our SR images being {\em perceptually} identical to the HR images.
Results on a large clinical dataset shows the benefits of each proposal, where our framework outperforms the state of the art quantitatively and
qualitatively.

\section {Methods}
\label {sec:methods}

We describe our novel framework, namely SRGAN\_SQE, through its architecture, loss functions, and learning algorithms.
Let the random-vector pair $( X^{\text{LR}}, X^{\text{HR}} )$ model the pair of a LR image patch $X^{\text{LR}}$ and its corresponding HR image patch
$X^{\text{HR}}$.
Let $P (X^{\text{LR}}, X^{\text{HR}})$ be their joint PDF modeling dependencies between the LR-HR patch pairs.
The training set has $N$ observed patch pairs $\{ (x^{\text{LR}}_i, x^{\text{HR}}_i) \}_{i=1}^N$, where each $(x^{\text{LR}}_i, x^{\text{HR}}_i)$ is
drawn independently from $P (X^{\text{LR}}, X^{\text{HR}})$.
%
%
In this paper, LR patches are sized 64$\times$64 pixels and HR (and SR) patches are sized 256$\times$256 pixels.

\subsection {Our SRGAN\_SQE Architecture}
\label {sec:arch}

\begin{figure}[!t]
  \twoAcrossLabelsHeight {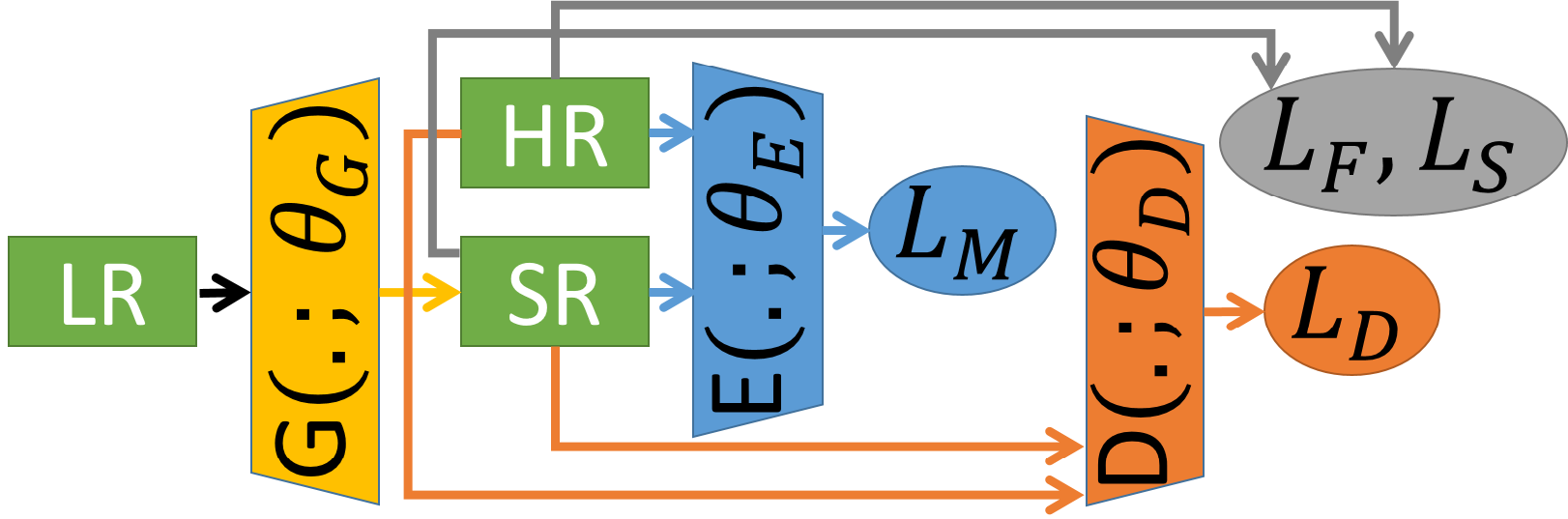} {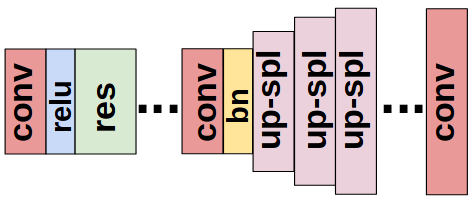} {{\bf (a)} Our SRGAN\_SQE Architecture} {{\bf (b)} Our Generator $\mathcal{G} (\cdot; \theta_G)$} {0.16}
  \twoAcrossLabelsHeight {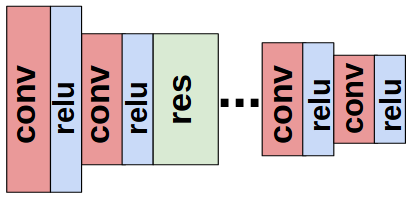} {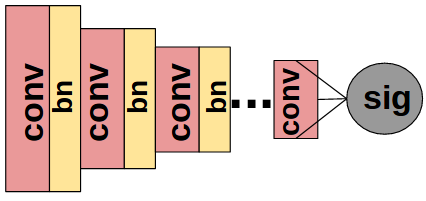} {{\bf (c)} Our Encoder $\mathcal{E} (\cdot; \theta_E)$} {{\bf (d)} Our Discriminator $\mathcal{D} (\cdot; \theta_D)$} {0.21}
  \vspace{-8pt}
  \caption
      {
        {\bf Our SRGAN\_SQE Framework.}
        {\bf (a)}~SRGAN\_SQE framework along with the architectures of its underlying components:
        {\bf (b)}~generator, {\bf (c)}~encoder, {\bf (d)}~discriminator.
      }
      \label{fig:arch}
\end{figure}

Our SRGAN\_SQE (Figure~\ref{fig:arch}(a)) comprises:
(i)~a {\em generator} $\mathcal{G} (\cdot; \theta_G)$ (Figure~\ref{fig:arch}(b)),
(ii)~an {\em encoder} $\mathcal{E} (\cdot; \theta_E)$ (Figure~\ref{fig:arch}(c)), and
(iii)~a {\em discriminator} $\mathcal{D} (\cdot; \theta_D)$ (Figure~\ref{fig:arch}(d)),
where $\theta_G$, $\theta_E$, and $\theta_D$ denote the associated trainable parameters.

{\bf Generator.}
Our generator learns a transformation function $\mathcal{G} (\cdot; \theta_G)$ on LR patches $X^{\text{LR}}$ to transform:
(i)~the PDF $P (X^{\text{LR}})$ into $P (X^{\text{HR}})$ and
(ii)~the patch $X^{\text{LR}}$ to the corresponding $X^{\text{HR}}$.
The associated loss function penalize the dissimilarity between the SR patch $\mathcal{G} (X^{\text{LR}}; \theta_G)$ and the corresponding HR patch
$X^{\text{HR}}$; detailed in Section~\ref{sec:loss}.
The generator (Figure~\ref{fig:arch}(b)) uses convolutional ({\em conv}) layers~\cite{cnn} with {\em relu} activation, residual ({\em res})
blocks~\cite{srgan}, batch-normalization ({\em bn}) layers~\cite{bn}, and up-sampling layers ({\em up-spl}).

{\bf Encoder.}
To penalize the dissimilarity between the SR patch $\mathcal{G} (X^{\text{LR}}; \theta_G)$ and the corresponding HR patch $X^{\text{HR}}$, we propose
a novel architecture to measure and penalize a robust {\em manifold-based distance} between the SR and HR patches; detailed in
Section~\ref{sec:loss}. We learn the manifold, specific to a class of images, in a pre-processing stage prior to performing super-resolution, by
training an autoencoder~\cite{baldi2012autoencoders} on HR patches $\{ x^{\text{HR}}_i \}_{i=1}^N$ in the training set. We design the autoencoder
using the encoder in Figure~\ref{fig:arch}(c) and an analogous mirror-symmetric decoder. After training the autoencoder, we use its encoder
$\mathcal{E} (\cdot; \theta_E)$ as a nonlinear mapper, to obtain the patches' manifold representations $\mathcal{E} (\mathcal{G} (X^{\text{LR}};
\theta_G); \theta_E)$ and $\mathcal{E} (X^{\text{HR}}; \theta_E)$. The manifold representation captures the texture characteristics in HR patches,
while reducing effects of noise and some artifacts.
The encoder (Figure~\ref{fig:arch}(c)) comprises {\em conv} layers, {\em relu} activation, and {\em res} blocks.

{\bf Discriminator.}
Our discriminator $\mathcal{D} (\cdot; \theta_D)$ learns to best differentiate the SR-patch PDF $P (\mathcal{G} (X^{\text{LR}}; \theta_G))$ and the
HR-patch PDF $P (X^{\text{HR}})$. Thus, for patches, say $y$, drawn from $P ( \mathcal{G} (X^{\text{LR}}; \theta_G) ) $ and $ P (X^{\text{HR}})$, the
discriminator output $\mathcal{D} (y; \theta_D)$ indicates the probability of $y$ belonging to the HR class. During training, the discriminator can
detect subtle differences between the PDFs of SR patches and (ground-truth) HR patches, to aid the generator to improve the matching between (the PDFs
of) SR patches and HR patches.
The discriminator (Figure~\ref{fig:arch}(d)) comprises {\em conv} layers, each followed by a {\em bn} layer, and use the sigmoid function to output
the class probabilities. The feature maps after every {\em conv} layer reduce the spatial dimensions by 2$\times$.
After training, the generator becomes very good at transforming LR patches close to their HR counterparts, thereby making the discriminator unable to
differentiate between the SR and HR patch PDFs, as desired.

\subsection {Our SRGAN\_SQE Formulation and Training}
\label {sec:loss}

Our SRGAN\_SQE optimizes $\theta_G$, $\theta_E$, and $\theta_D$ using novel loss functions designed to
(i)~be robust to training-set corruptions and
(ii)~model realistic texture and human perception.

{\bf Robust Fidelity Loss.}
Typical GAN models penalize the MSE between the SR image $\mathcal{G} (X^{\text{LR}}; \theta_G)$ and the corresponding HR image $X^{\text{HR}}$. The
MSE loss assumes the residual $\mathcal{G} (X^{\text{LR}}; \theta_G) - X^{\text{HR}}$ to have an isotropic Gaussian PDF, which can be incorrect
because of training set corruptions due to errors in specimen preparation and staining, imaging artifacts, or human errors in dataset curation.
For corrupted training sets, residual PDFs can have significantly heavier tails than a Gaussian, where the tails comprise examples that are clinically
irrelevant. Such corruptions force the network to additionally learn mappings for LR-HR patch pairs that are very {\em atypical}, or even {\em
  outliers}, thereby risking a significant reduction in the performance of the network on patches that are actually {\em clinically relevant} for the
application.
Indeed, results in Section~\ref{sec:results} support this claim.
Thus, we propose to minimize the $q$-th power of the $q$-quasi-norm of the residual vector, where $q \in (0,1)$, thereby modeling the residual PDF as
{\em generalized Gaussian}~\cite{novey2010complex}\cite{meet_paper}, with smaller $q$ leading to heavier tails.
We make the loss function differentiable and usable in back-propagation, we use the $\epsilon$-regularised quasi norm.
So, our novel robust fidelity loss is $ \mathcal{L}_F (\theta_G) := E_{P (X^{\text{LR}}, X^{\text{HR}})} [ \| \mathcal{G} (X^{\text{LR}}; \theta_G) -
  X^{\text{HR}} \|_{q,\epsilon}^q ] $ , where, for any image patch $a$ with pixel values $a_i$, we design $\| a \|_{q,\epsilon}^q := \sum_i (|a_i|^2 +
\epsilon)^{q/2}$, with $\epsilon$ as small positive real found empirically.
Our robust loss reweights of the MSE-based gradient that is proportional to the residual, such that gradients associated with
larger residuals (e.g., caused by tail data or outliers) get scaled down for stable learning.

{\bf Robust Manifold-Based Loss.}
Our robust fidelity loss $\mathcal{L}_F$ enforces an independent pixelwise matching of the SR patch $\mathcal{G} (X^{\text{LR}}; \theta_G)$ with the
corresponding HR patch $X^{\text{HR}}$.
In addition, we also propose to penalize the dissimilarity in the underlying textural structures between $\mathcal{G} (X^{\text{LR}}; \theta_G)$ and
$X^{\text{HR}}$, to capture the dependencies of pixel values across the entire patch. Hence, we propose a novel loss function that captures this
dissimilarity by (pre-)learning the manifold representation of HR images through the encoder $\mathcal{E} (\cdot; \theta_E)$ described in
Section~\ref{sec:arch}.
In this manifold representation as well, the PDF of the residuals can be heavy tailed and, thus, we propose a novel robust loss relying on the
manifold distance between SR patches and the corresponding HR patches: $ \mathcal{L}_M (\theta_G, \theta_E) := E_{P (X^{\text{LR}}, X^{\text{HR}})} [
  \| \mathcal{E} (\mathcal{G} (X^{\text{LR}}; \theta_G); \theta_E) - \mathcal{E} (X^{\text{HR}}; \theta_E) \|_{q,\epsilon}^q ] $ .

{\bf Perception-Based Loss.}
In clinical applications, the SR image is interpreted by a human expert, e.g., a pathologist. Thus, we want SR patches to be {\em perceptually}
identical to the (ground-truth) HR patches.
During training, we propose to enforce human-perceptual similarity between the SR and HR patches by penalizing the negative sum of structural
similarity (sSSIM)~\cite{ssim} values between SR and HR patches (sum taken over all overlapping $5 \times 5$ neighborhoods in the SR and HR
patches). We compute sSSIM on each of the three color channels (R,G,B) and add them up to give the overall sSSIM. This sSSIM accounts for perceptual
changes in structural information, luminance, and contrast over all neighborhoods between the SR and HR patches. Thus, we propose the perception-based
loss as $ \mathcal{L}_S (\theta_G) := - E_{P (X^{\text{LR}}, X^{\text{HR}})} [ \text{sSSIM} (\mathcal{G} (X^{\text{LR}}; \theta_G), X^{\text{HR}}) ] $
.

{\bf Discriminator Loss.}
Our discriminator training penalizes the Kullback-Leibler (KL) divergences between
(i)~the probability vectors (distributions) for the SR and HR patches produced by the discriminator, i.e., $[ \mathcal{D} (\mathcal{G} (X^{\text{LR}};
  \theta_G); \theta_D), 1-\mathcal{D} (\mathcal{G} (X^{\text{LR}}; \theta_G); \theta_D) ]$ or $[ \mathcal{D} ( X^{\text{HR}}; \theta_D), 1-\mathcal{D}
  (X^{\text{HR}}; \theta_D) ]$, and
(ii)~the one-hot probability vectors (distributions) for the SR and HR patches, i.e., $[0, 1]$ or $[1, 0]$, respectively.
Thus, our discriminator-based loss function is $ \mathcal{L}_D (\theta_G, \theta_D) := E_{P (X^{\text{LR}}, X^{\text{HR}})} [ \log (1-\mathcal{D}
  (\mathcal{G} (X^{\text{LR}}; \theta_G); \theta_D)) + \log (\mathcal{D} ( X^{\text{HR}}; \theta_D)) ] $ .
On one hand, the generator optimizes $\theta_G$ to reduce the loss function $\mathcal{L}_D$, where the generator learns to produce high-quality SR
patches $\mathcal{G} (X^{\text{LR}}; \theta_G)$ that ``trick'' the discriminator in assigning them a high probability $\mathcal{D} (\mathcal{G}
(X^{\text{LR}}; \theta_G); \theta_D)$ of being in the HR class, i.e., being drawn from $P (X^{\text{HR}})$.
On the other hand, the discriminator training optimizes $\theta_D$ to increase the ``gain'' function $\mathcal{L}_D$, where the discriminator learns
to give
(i)~HR patches high probabilities $\mathcal{D} ( X^{\text{HR}}; \theta_D)$ of belonging to the HR class and
(ii)~SR patches low probabilities $\mathcal{D} (\mathcal{G} (X^{\text{LR}}; \theta_G); \theta_D)$ of belonging to the HR class.

{\bf SRGAN\_SQE Formulation and Training.}
We train our SRGAN\_SQE to optimize the parameters $\theta_G, \theta_D$ by solving $\arg \min_{\theta_G} \arg \max_{\theta_D} [
  \mathcal{L}_F (\theta_G) + \lambda_M \mathcal{L}_M (\theta_G, \theta_E) + \lambda_S \mathcal{L}_S (\theta_G) + \lambda_D \mathcal{L}_D
  (\theta_G, \theta_D) ]$, where $\lambda_M, \lambda_S, \lambda_D$ are positive real free parameters that control the balance across the
loss functions.
In this paper, $\lambda_M := 0.2$, $\lambda_S := 2$, $\lambda_D := 0.016$.
We use alternating optimization on $\theta_G$ and $\theta_D$, using back-propagation and Adam optimizer~\cite{adam} (for all $q$).
We train the generator using the non-saturating heuristic~\cite{gan}.

\section {Results and Discussion}
\label {sec:results}

We evaluate using histopathology images from the CAMELYON dataset~\cite{camelyon}. We extract 4000 patches of size $256 \times 256$ pixels at the
highest resolution and consider them as HR ground truth. We Gaussian-smooth and subsample HR patches to create LR patches ($64 \times 64$ pixels). We
use 2000 patches for training and 2000 for testing.
We evaluate four methods:
(i)~SRGAN~\cite{srgan}, a current state of the art;
(ii)~our SRGAN\_E, where $q := 2, \lambda_S := 0$;
(iii)~our SRGAN\_QE, where $q \in (0,1), \lambda_S := 0$;
(iv)~our SRGAN\_SQE, where $q \in (0,1), \lambda_S > 0$.
We evaluate performance using three quantitative measures:
(i)~relative root MSE (RRMSE) between SR and HR images, i.e., $\| \mathcal{G} (X^{\text{LR}}; \theta_G) - X^{\text{HR}} \|_{\text{F}} / \|
X^{\text{HR}} \|_{\text{F}}$,
(ii)~multiscale mean SSIM (MS-mSSIM) that averages SSIM across neighborhoods at five different scales~\cite{ssim},
(iii)~quality index based on local variance (QILV)~\cite{qilv}.
While MS-mSSIM is more sensitive to random noise than blur in the images, QILV acts complementarily and is more sensitive to blur than noise.

\begin{figure}[!t]
  \threeAcrossLabelsHeight {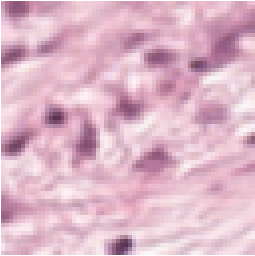}   {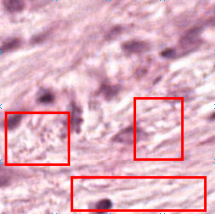}    {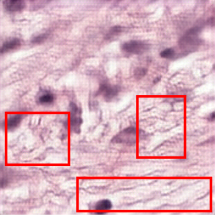}    {{\bf (a1)} Low-res.} {{\bf (b1)} Ground truth} {{\bf (c1)} \textcolor{orange}{SRGAN}} {0.32}
  \threeAcrossLabelsHeight {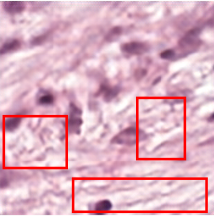} {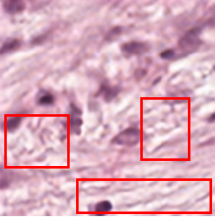} {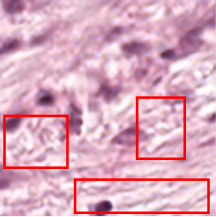} {{\bf (d1)} \textcolor{blue}{SRGAN\_E}} {{\bf (e1)} \textcolor{green}{SRGAN\_QE}} {{\bf (f1)} \textcolor{red}{SRGAN\_SQE}} {0.32}
  \threeAcrossLabelsHeight {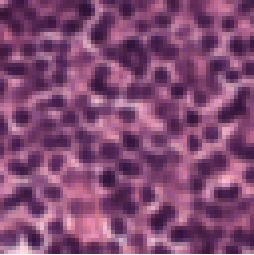}   {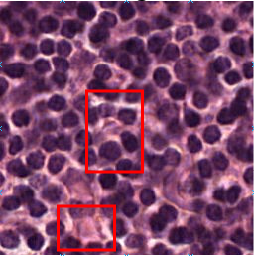}    {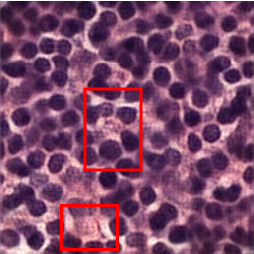}    {{\bf (a2)} Low-res.} {{\bf (b2)} Ground truth} {{\bf (c2)} \textcolor{orange}{SRGAN}} {0.32}
  \threeAcrossLabelsHeight {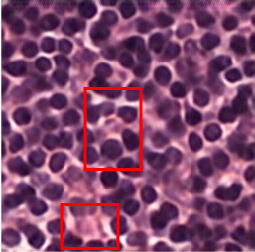} {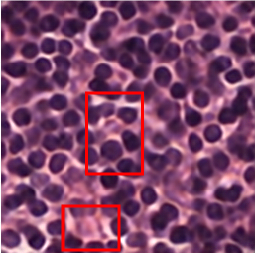} {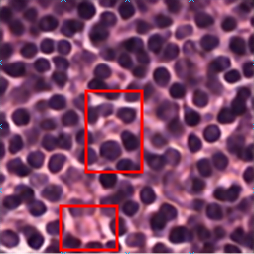} {{\bf (d2)} \textcolor{blue}{SRGAN\_E}} {{\bf (e2)} \textcolor{green}{SRGAN\_QE}} {{\bf (f2)} \textcolor{red}{SRGAN\_SQE}} {0.32}
  \vspace{-8pt}
  \caption
      {
        {\bf Results without Explicitly Corrupting Training Set.}
        {\bf (a1)-(a2)}~LR input.
        {\bf (b1)-(b2)}~HR ground truth.
        [MS-mSSIM,RRMSE,QILV] for
        {\bf (c1)-(c2)}~SRGAN: \textcolor{orange}{[0.67,0.004,0.973]}, \textcolor{orange}{[0.87,0.14,0.972]};
        {\bf (d1)-(d2)}~SRGAN\_E: \textcolor{blue}{[0.86,0.001,0.975]}, \textcolor{blue}{[0.87,0.011,0.973]};
        {\bf (e1)-(e2)}~SRGAN\_QE: \textcolor{green}{[0.89,0.001,0.987]}, \textcolor{green}{[0.89,0.007,0.984]};
        {\bf (f1)-(f2)}~{\bf SRGAN\_SQE}: \textcolor{red}{[0.91,0.001,0.996]}, \textcolor{red}{[0.90,0.008,0.993]}.
      }
      \label{fig:results}
\end{figure}

\begin{figure}[!t]
  \twoAcrossLabelsHeight {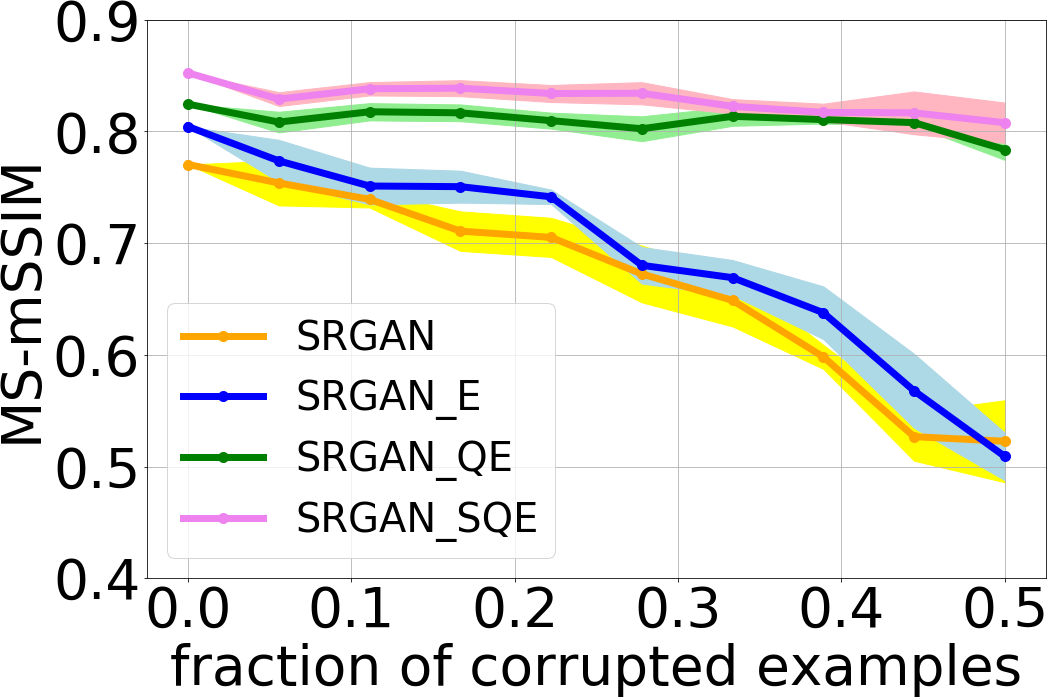}  {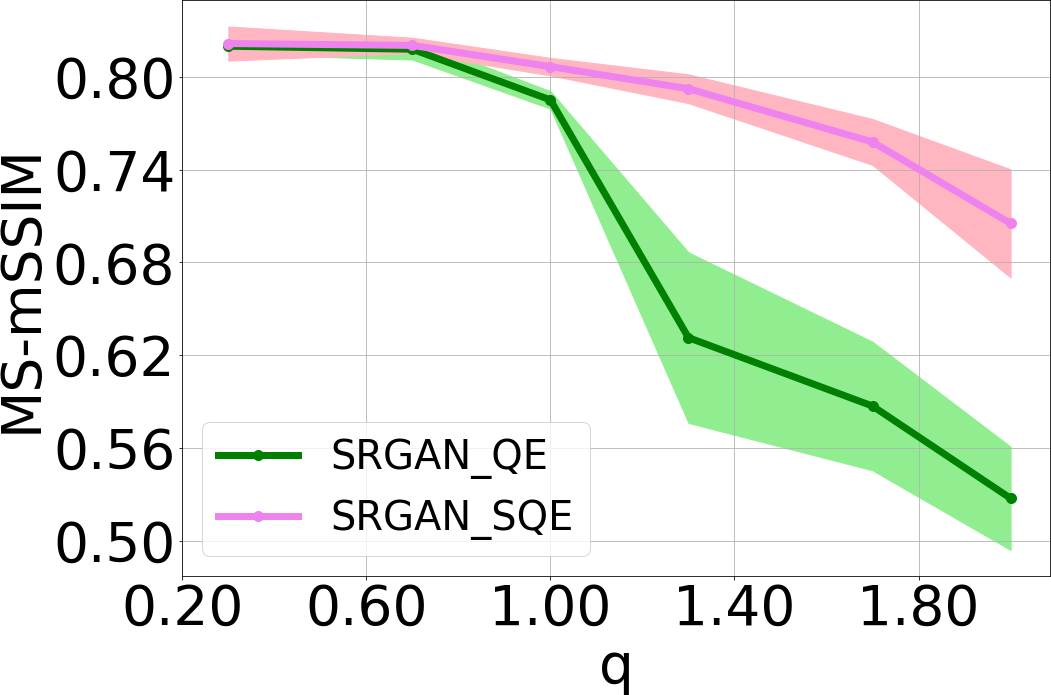}  {\bf (a1)} {\bf (b1)} {0.32}
  \vspace{-2pt}
  \twoAcrossLabelsHeight {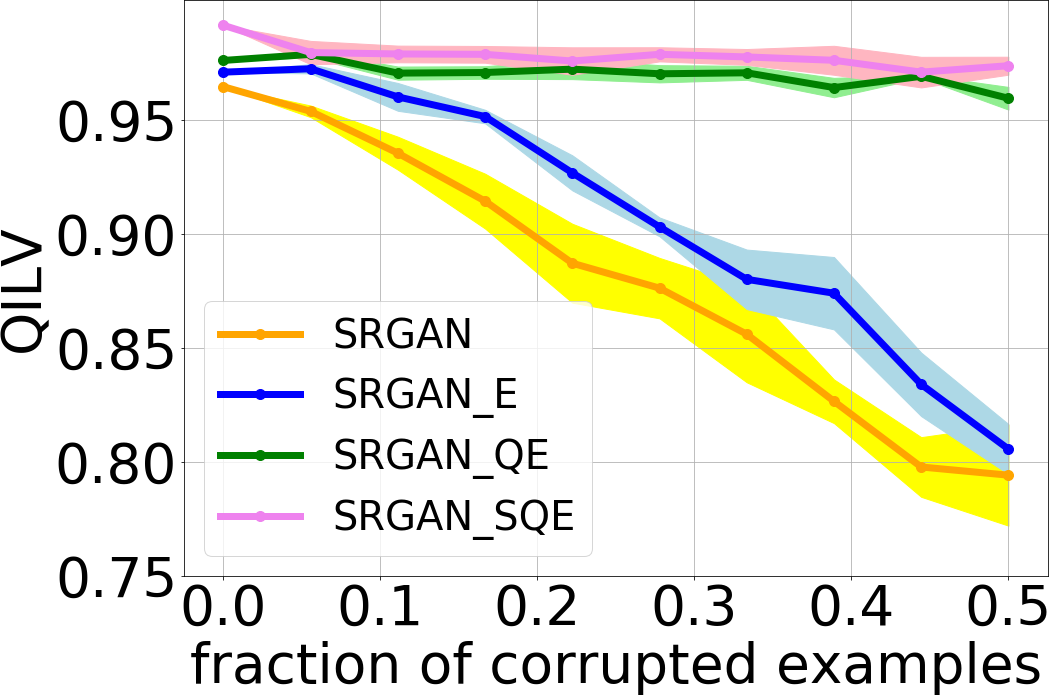}  {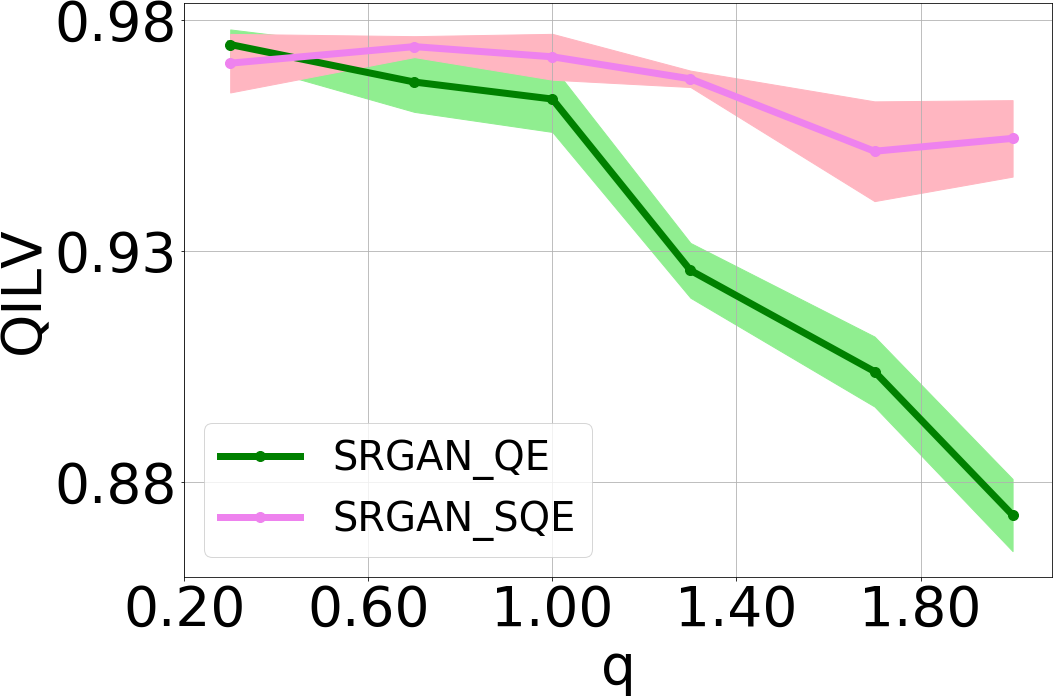} {\bf (a3)} {\bf (b3)} {0.32}
  \vspace{-2pt}
  \twoAcrossLabelsHeight {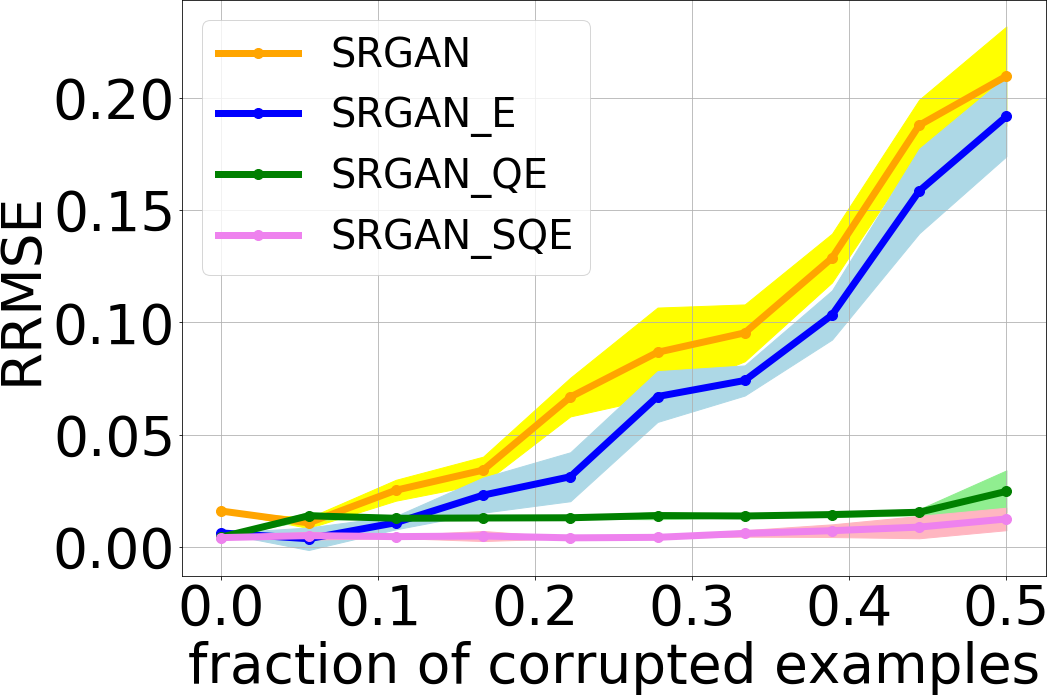} {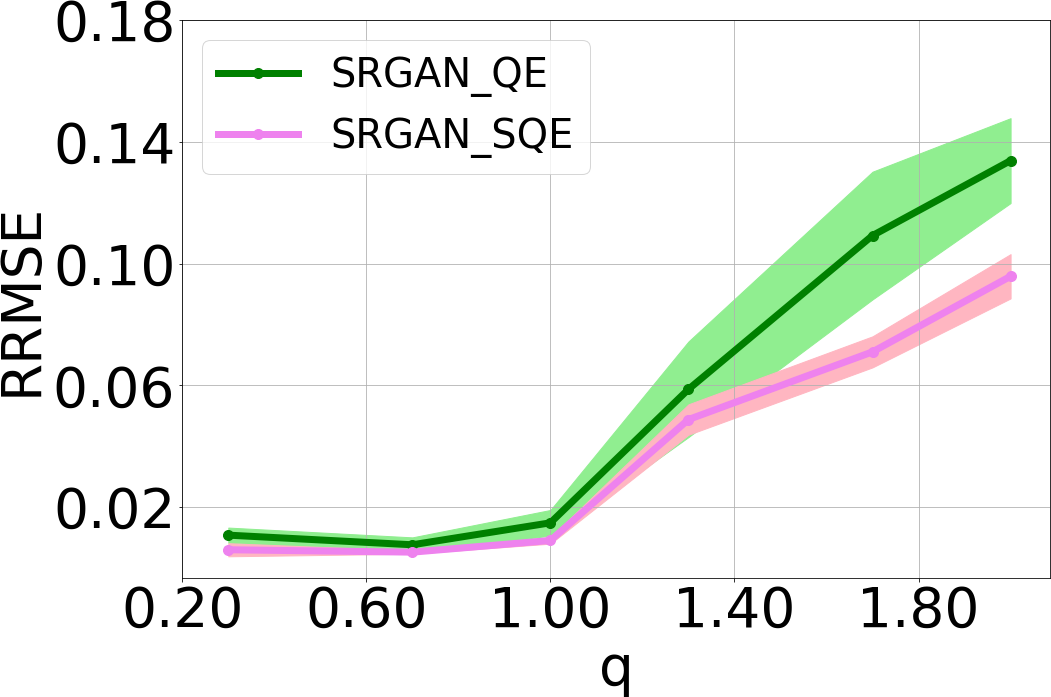} {\bf (a2)} {\bf (b2)} {0.32}
  \vspace{-8pt}
  \caption
      {
        {\bf Results with Varying Levels of Training-set Corruption and Robustness $q$.}
        Plots for MS-mSSIM, QILV, and RRMSE for:
        {\bf (a1)--(a3)}~varying fractions of corrupted examples introduced in training set ($q$=0.5);
        {\bf (b1)--(b3)}~varying robustness $q$ (corrupted fraction in training dataset =$0.3$).
      }
      \label{fig:graphs}
\end{figure}

We start by evaluating methods on our training set that is well curated and does {\em not} contain any explicitly-introduced corrupted examples. The
training data exhibit {\em natural} variability in terms of texture, contrast, intensity histograms, and focus.
First, replacing the VGG-encoding in SRGAN (Figure~\ref{fig:results}(c1)-(c2)) with our (pre-)learned autoencoder-based encoding $\mathcal{E}
(\cdot;\theta_E)$ and manifold-based loss $\mathcal{L}_M$ in SRGAN\_E improves the results qualitatively (Figure~\ref{fig:results}(d1)-(d2)) and
quantitatively (Figure~\ref{fig:graphs}(a1)--(a3)).
Second, with $q < 1$, our SRGAN\_QE (Figure~\ref{fig:results}(e1)-(e2)), improves over SRGAN\_E, indicating that the distribution of residuals (in
spatial and encoded domains) even with {\em natural} variability is much heavier tailed than Gaussian (best results for $q \approx 0.5$). The
improvement is evident quantitatively in Figure~\ref{fig:graphs}(a1)--(a3) when fraction of corrupted examples is zero.
Our perceptual sSSIM-based loss $\mathcal{L}_S$, even at a single scale of $5 \times 5$ pixel neighborhoods, in our SRGAN\_SQE leads to further
improvements, when our SR images (Figure~\ref{fig:results}(f1)-(f2)) become virtually identical to the HR image qualitatively
(Figure~\ref{fig:results}(b1)-(b2)) and quantitatively (Figure~\ref{fig:graphs}(a1)--(a3)).

\begin{figure}[!t]
  \threeAcrossLabelsHeight {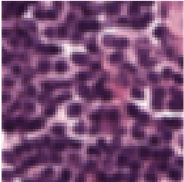} {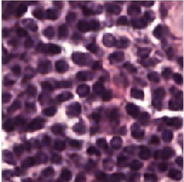} {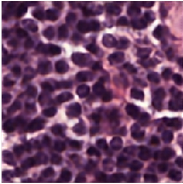} {{\bf (a)} Low-res. input} {{\bf (c1)} \textcolor{red}{SRGAN\_SQE}} {{\bf (d1)} \textcolor{orange}{SRGAN}} {0.315}
  \threeAcrossLabelsHeight {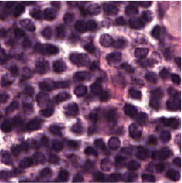} {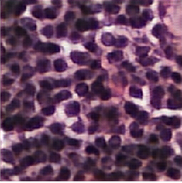}  {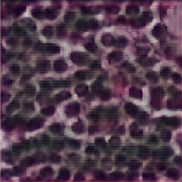}  {{\bf (b)} Ground truth}   {{\bf (c2)} \textcolor{red}{SRGAN\_SQE}} {{\bf (d2)} \textcolor{orange}{SRGAN}} {0.315}
  \vspace{-8pt}
  \caption
      {
        {\bf Results with Varying Levels of Training-set Corruption.}
        {\bf (a)}~LR input.
        {\bf (b)}~HR ground truth.
        Results for {\bf (c1)-(c2)}~{\bf Our SRGAN\_SQE} and {\bf (d1)-(d2)}~SRGAN,
        with $5\%$ and $30\%$ corrupted examples introduced in the training set.
        [MS-mSSIM,RRMSE,QILV] for: (c1)-(c2)~{\bf Our SRGAN\_SQE} are \textcolor{red}{[0.91,0.01,0.993]} and \textcolor{red}{[0.91,0.01,0.987]};
        (d1)-(d2) SRGAN are \textcolor{orange}{[0.85,0.02,0.976]} and \textcolor{orange}{[0.76,0.15,0.853]}.
      }
      \label{fig:resultsCorr}
\end{figure}

Next, we evaluate all methods by explicitly introducing corrupted examples during training, in the form of increasing levels of noise (e.g., from weak
signals), blur (e.g., from poor focus), or changes in contrast (e.g., from poor staining).
Quantitatively, our SRGAN\_SQE outperforms all other methods, for varying levels of corruption from 0\% to 50\% (Figure~\ref{fig:graphs}(a1)--(a3)).
Modeling heavy-tailed residuals with $q \approx 0.5$ gives the best results across all measures (Figure~\ref{fig:graphs}(b1)--(b3)).
As the level of corruptions increase from 5\% to 30\%, the results from our SRGAN\_SQE (Figure~\ref{fig:resultsCorr}(c1)-(c2)) stay stable and
high-quality, visually and quantitatively, but the results from SRGAN (Figure~\ref{fig:resultsCorr}(d1)-(d2)) degrade significantly.

{\bf Conclusion.}
We proposed the novel SRGAN\_SQE framework for super-resolution that makes learning {\em robust} to errors in training-set curation by modeling {\em
  heavy-tailed} PDFs, using {\em quasi norms}, on the residuals in the spatial and {\em manifold-encoding} domains. We learned the manifold, using an
autoencoder, to reproduce realistic {\em textural} details.
We proposed a sSSIM based loss, at a single fine scale, to output SR images matching {\em human perception}.
Results on a large clinical dataset showed that our SRGAN\_SQE improves the quality of the super-resolved images over the state of the art
quantitatively and qualitatively.

\bibliographystyle{IEEEbib}
\bibliography{./Bibtex_SRGAN.bib}

\end{document}